\documentclass[twoside,11pt]{article}

\pdfoutput=1

\usepackage{bm}
\usepackage{amsmath}
\usepackage{amsfonts}
\usepackage{mathtools}
\usepackage{comment}
\usepackage{subfigure}
\usepackage{marvosym}

\newcommand{\tp}{^{\top}}

\DeclareMathOperator*{\argmax}{arg\,max}
\DeclareMathOperator{\KLt}{KL}
\DeclarePairedDelimiterX{\infdivx}[2]{[}{]}{%
  #1\;\delimsize\|\;#2%
}
\newcommand{\KL}{\KLt\infdivx}
\DeclareMathOperator{\bernoullipdf}{Bernoulli}
\DeclareMathOperator{\normalpdf}{N}
\DeclareMathOperator{\lognormalpdf}{Log-N}
\DeclareMathOperator{\gammapdf}{Gamma}
\DeclareMathOperator{\E}{\mathbb{E}}
\DeclareMathOperator{\betapdf}{Beta}

\begin{document}

\title{Improving drug sensitivity predictions in precision medicine through active expert knowledge elicitation}

\author{Iiris Sundin\,$^{\text{1,}*}$, Tomi Peltola\,$^{\text{1}}$, Muntasir Mamun Majumder\,$^{\text{2}}$, \\Pedram Daee\,$^{\text{1}}$, Marta Soare\,$^{\text{1}}$, Homayun Afrabandpey\,$^{\text{1}}$, \\ Caroline Heckman\,$^{\text{2}}$, Samuel Kaski\,$^{\text{1,\Cross,}*}$ and Pekka Marttinen\,$^{\text{1,\Cross,}*}$}

\date{}
\maketitle

\vspace{-1cm}
\begin{center}
\small $^{\text{\sf 1}}$Helsinki Institute for Information Technology HIIT, Department of Computer Science, Aalto University, Finland \\
$^{\text{\sf 2}}$Institute for Molecular Medicine Finland (FIMM), University of Helsinki, Finland.

$^\ast$To whom correspondence should be addressed.

$^\text{\Cross}$These authors contributed equally to this work.
\end{center}

\begin{abstract}
Predicting the efficacy of a drug for a given individual, using high-dimensional genomic measurements, is at the core of precision medicine. However, identifying features on which to base the predictions remains a challenge, especially when the sample size is small. Incorporating expert knowledge offers a promising alternative to improve a prediction model, but collecting such knowledge is laborious to the expert if the number of candidate features is very large. We introduce a probabilistic model that can incorporate expert feedback about the impact of genomic measurements on the sensitivity of a cancer cell for a given drug. We also present two methods to intelligently collect this feedback from the expert, using experimental design and multi-armed bandit models. In a multiple myeloma blood cancer data set ($n$=51), expert knowledge decreased the prediction error by 8\%. Furthermore, the intelligent approaches can be used to reduce the workload of feedback collection to less than 30\% on average compared to a naive approach.
\end{abstract}

\section{Introduction}
In genomics-based personalized or precision medicine, large-scale screenings and sequencing produce thousands of genomic and molecular features for each sample. However, the data sets are small: typically only hundreds, or at most a thousand cell lines, and even fewer patients, are included in the data sets. 
This is the case 
also for genomic cancer medicine, the focus of this work, which includes gene expression, somatic mutation, copy number variation and cytogenetic marker measurements characterizing \emph{ex vivo} bone marrow patient samples for the task of predicting sensitivity to a panel of drugs. Although large-scale genomic studies in cancer have identified recurrent molecular events that predict prognosis and explain pathogenesis, causal effects on drug response have been established for only a few of those features. The limited data poses a challenge for learning predictive models from the data. The established statistical methods for finding predictive features (biomarkers) and learning predictive models are similar across omics-based data analysis tasks. Multivariate analysis of variance \cite{garnett2012systematic} is a classical linear method. Recently, sparse regression models such as LASSO and elastic net \cite{garnett2012systematic, jang2014systematic} have become standard reliable benchmark methods, and kernel methods enable finding more complex nonlinear combinations of the features \cite{costello2014community,muhammad2016drug}.

\begin{figure*}[h]
\centering
\includegraphics[width=0.75\textwidth,keepaspectratio]{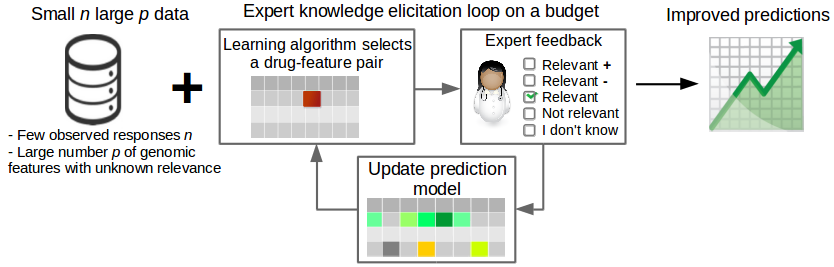}
\caption{Overview. Predictions in small sample size problems are improved by asking experts in an elicitation loop. The system presents questions for the expert sequentially to maximize performance with a minimal number of questions, i.e., on a budget. The expert answers the questions by indicating whether a feature is relevant and in which direction in predicting the response to a drug.}
\label{fig:overview}
\end{figure*}

A natural way to solve the particular problems caused by a small sample size is to measure more data. This is, however, often not an available option, due to costs, risks, or the rarity of the disease. Statistical ways of alleviating the problem are multitask learning~\cite{yuan2016multitask, muhammad2016drug}, which increases predictive power by sharing statistical strength between multiple related outputs or data sets, and incorporating biological prior knowledge. Biological prior knowledge about cancer pathways has been used as side information for learning~\cite{costello2014community, muhammad2016drug}, for feature selection~\cite{jang2015stepwise, de2016algorithms}, or to modify regularization of, for instance, an elastic net~\cite{sokolov2016pathway}. Although these methods improve the predictions, they sidestep the problem of what information to choose from the databanks, and naturally cannot include knowledge not yet in the databanks.

A more rarely exploited alternative is to ask an expert. \textit{Prior elicitation techniques} \cite{OHagan06} have been used for constructing prior distributions for Bayesian data analysis that take into account expert knowledge, and hence can restrict the range of parameters to be later used in learning models~\cite{garthwaite1988quantifying,garthwaite2013prior,kadane1980interactive, afrabandpey2016interactive}. These techniques focus on how to reliably elicit knowledge, whereas in practice it is equally important to minimize the effort required from the expert. \emph{Interactive and sequential learning} can help by carefully deciding what to ask the user, and has been used, for instance, for clustering~\cite{must-link-cannot-link,Balcan2008}, learning of Bayesian networks~\cite{Cano2011a}, and for visualization~\cite{House2015bayesian}. 

Very recently, interactive learning has been proposed for including expert knowledge in a prediction task, in a linear regression setting with a small sample size. First indications that improvements are possible~\cite{soare16regression} were obtained with strong assumptions on simulated experts. Human experts were included in two other studies~\cite{Daee2016,Micallef2017} on textual data, obtaining improved predictions with a small number of expert feedbacks, for the tasks of predicting user ratings, and predicting citation counts. The elicitation techniques were based on Bayesian experimental design and a multi-armed bandit user model, which helped the user solve the exploration-exploitation dilemma~\cite{auer2002using} in giving feedback.

In this paper, we introduce sequential knowledge elicitation methods to the precision medicine prediction task, illustrated in Figure~\ref{fig:overview}. 
As a case study to show the potential of the methods, we use drug sensitivity prediction in patient samples, which is known to be very hard. We predict drug responses of \emph{ex vivo} cell samples from blood cancer patients, based on mutation data and cytogenetics markers. Two well-informed experts are asked to provide feedback about the relevance of features when predicting sensitivity to specific targeted drugs, and about the direction of the putative effects.

Our main contribution is to show for the first time that sequential expert knowledge elicitation can improve predictive modeling in precision medicine. Specifically, we show that
\begin{itemize}
    \item expert knowledge elicitation improves the accuracy of drug sensitivity predictions in the difficult case of predicting drug responses based on patient's somatic mutations and cytogenetic markers, crucial for choosing which drug to prescribe to a patient in precision medicine,
    \item sequential expert knowledge elicitation reduces the number of queries required from the expert compared to naive approach with randomly chosen queries.
\end{itemize}
These empirical results required four advances from recent \cite{Daee2016,Micallef2017} methods proposed for different elicitation tasks:
Firstly, we extend a method previously used in automatic design of experiments to the challenging expert knowledge elicitation task in precision medicine. In addition, we introduce feedback on the direction of the putative effect and show that it is more effective in improving the drug sensitivity predictions than general relevance feedback. We develop the bandit user model approach to incorporate biological information in the form of pathways and drug targets. Finally, we extend the methods from univariate to multivariate outputs (sensitivity to multiple drugs, with feedback given to (drug,~feature) pairs).

\section{Problem set-up}
In cancer treatment, doctors need to choose which of the available drugs to administer to new patients. Machine learning could be used to assist the doctor in the choice by predicting the drug responses of patients from their genomic features. The available patient data sets, however, are too small for accurate learning of predictive models. Here we aim at improving the drug response predictions in this challenging setting. In particular, we propose sequential expert knowledge elicitation as a solution to the problem of drug response prediction given a small sample size. In our specific case, data from 51 blood cancer patients are available, with the number of considered genomic features (mutations and cytogenetic markers) being 3032, and we wish to predict the responses for 12 drugs. 

Experts in blood cancer medicine have knowledge of biomarkers, 
and could also associate other features to the drug responses based on their experience. Unfortunately, the approach of naively querying each feature for knowledge from the experts is burdensome given the large number of features. Therefore, we introduce two sequential knowledge elicitation algorithms that are able to choose the (drug,~feature) pairs that have the highest effect in improving the predictions, and compare their performance in this precision medicine case.

We assume that the experts will be able to answer two types of questions regarding the effect of a feature to drug response. Firstly, if a feature is \textit{relevant} (or \textit{irrelevant}) in predicting the drug response. This type of feedback was used in~\cite{Daee2016} for textual data. 
In addition to that, the experts may possess knowledge on the direction of the effect for a subset of relevant features. This \textit{directional} feedback tells if a feature is positively or negatively correlated with the drug response, extending the work in~\cite{Micallef2017}, where the feedback was only on positive correlations and used in a textual data application. 
The precise mathematical formulation of the effect of the two feedback types to the prediction model will be given in Section~\ref{Prediction model}.

\section{Models and algorithms}
In this section, we describe the proposed models and algorithms for sequential expert knowledge elicitation. First, we describe a sparse linear regression model that is used to learn the relationship between genomic features and drug responses, and which takes into account the elicited expert knowledge. Then we introduce the two elicitation methods developed for the case study in precision medicine.

\subsection{Prediction model} \label{Prediction model}
Sparse linear regression models are used to predict the drug sensitivities based on the genomic features. Let $y_{n,d}$ be the sensitivity of the $n$th patient for drug $d$, and $\bm{x}_n \in \mathbb{R}^M$ be the vector of the patient's $M$ genomic features. We assume a Gaussian observation model:
\begin{equation*}
y_{n,d} \sim \normalpdf(\bm{w}_d\tp \bm{x}_n, \sigma^2_d),
\end{equation*}
where the $\bm{w}_d \in \mathbb{R}^M$ are the regression weights and $\sigma^2_d$ is the residual variance. A sparsity-inducing spike-and-slab prior \cite{mitchell1988bayesian, george1993variable} is put on the weights:
\begin{equation*}
w_{d,m} \sim \gamma_{d,m} \normalpdf(0, \tau_{d,m}^2) + (1 - \gamma_{d,m}) \delta_0,
\end{equation*}
where $\gamma_{d,m}$ is a binary variable indicating whether the $m$th feature is relevant (i.e., $w_{d,m}$ drawn from a zero-mean Gaussian prior with variance $\tau_{d,m}^2$) or not ($w_{d,m}$ is set to zero via the Dirac delta spike $\delta_0$) when predicting for the $d$th drug. The prior probability of relevance $\rho_d$ controls the expected sparsity of the model via the prior
\begin{equation*}
\gamma_{d,m} \sim \bernoullipdf(\rho_d).
\end{equation*}

The model is completed with the hyperpriors:
\begin{align*}
    \sigma^{-2}_d &\sim \gammapdf(\alpha_{\sigma}, \beta_{\sigma}),\\
    \rho_d &\sim \betapdf(\alpha_\rho, \beta_\rho),\\
    \tau_{d,m} &\sim \lognormalpdf(\mu, \omega^2).
\end{align*}
Settings for the values of the hyperparameters are discussed in Section~\ref{sec:exp_methods}.

Expert knowledge is incorporated into the model via feedback observation models \cite{Daee2016}. The relevance feedback $f^{rel}_{d,m} \in \{0,1\}$ ($0$ denotes irrelevant, $1$ relevant) of feature $m$ for drug $d$ follows
\begin{equation*}
f^{rel}_{d,m} \sim \gamma_{d,m} \bernoullipdf(\pi_d^{rel}) + (1 - \gamma_{d,m}) \bernoullipdf(1 - \pi_d^{rel}),
\end{equation*}
where $\pi_d^{rel}$ is the probability of the expert being correct. For example, when the $m$th feature for drug $d$ is relevant in the regression model (i.e., $\gamma_{d,m} = 1$), the expert would \textit{a priori} be assumed to say $f^{rel}_{d,m} = 1$ with probability $\pi_d^{rel}$. In the model learning
, once the expert has provided the feedback based on his or her knowledge, $\pi_d^{rel}$ effectively controls how strongly the model will change to reflect the feedback.

The directional feedback $f^{dir}_{d,m} \in \{0,1\}$ ($0$ denotes negative weight, $1$ positive) follows
\begin{equation*}
f^{dir}_{d,m}\!\sim\!I(w_{d,m}\!\geq\!0)\!\bernoullipdf(\pi_d^{dir})\!+\!I(w_{d,m}\!<\!0)\!\bernoullipdf(1\!-\!\pi_d^{dir}),
\end{equation*}
where $I(C) = 1$ when the condition $C$ holds and $0$ otherwise, and $\pi_d^{dir}$ is again the probability of the expert being correct. For example, when the weight $w_{d,m}$ is positive, the expert would \textit{a priori} be assumed to say $f^{dir}_{d,m} = 1$ with probability $\pi_d^{dir}$. To simplify the model, we assume $\pi_d = \pi_d^{dir} = \pi_d^{rel}$ and set a prior on $\pi_d$ as $\pi_d \sim \betapdf(\alpha_\pi, \beta_\pi)$. The prediction model and learning are detailed in Appendix A.

\subsection{Expert knowledge elicitation methods}
The purpose of expert knowledge elicitation algorithms is to sequentially choose queries to the expert, so that the improvement in predictions is maximized. In the case of genomic data, very few genes are known to be associated with drug responses, and therefore restricting the suggestions to the possibly known genes greatly reduces the number of queries. Previously, two expert knowledge elicitation methods have shown promising results for prediction of a single outcome variable on textual data. We extend these methods to the multi-output precision medicine setting. Next, we will introduce the two alternative elicitation methods, the performances of which will be evaluated and compared in the experiments.

\subsubsection{Sequential experimental design} \label{information_gain}

We introduce a sequential experimental design approach to select the next (drug,~feature) pair candidate to be queried for feedback from the expert, extending the work in \cite{Daee2016}. Specifically, at each iteration $t$, we find the pair where the feedback from the expert is expected to have the maximal influence on the drug sensitivity prediction. The amount of information in the expert feedback is measured by the Kullback--Leibler divergence ($\KLt$) between the predictive distributions before and after observing the feedback. As the feedback value itself is unobserved before the actual query, an expectation over the predictive distributions of the two types of feedbacks is computed in finding the (drug,~feature) pair $(d^*,m^*)$ with the highest expected information gain:
\begin{align*}
&\argmax_{(d,m)\notin \bm{F}_{t-1}}\E_{\tilde{f}^{rel}_{d,m},\tilde{f}^{dir}_{d,m} \mid \mathcal{D}_{t-1}}\left[\sum_{n=1}^N u_{n,d,m,t}\right] \text{ where}\\
& u_{n,d,m,t} = \KL{p(\tilde{y}_d \mid \bm{x}_n, \mathcal{D}_{t-1}, \tilde{f}^{rel}_{d,m}, \tilde{f}^{dir}_{d,m}) }{ p(\tilde{y}_d \mid \bm{x}_n,\mathcal{D}_{t-1})},
\end{align*}
and $\mathcal{D}_{t-1} = (\bm{Y}, \bm{X}, \bm{F}_{t-1})$, observed drug sensitivities $\bm{Y} \in \mathbb{R}^{N \times D}$ for $N$ patients and $D$ drugs and the genomic features $\bm{X} \in \mathbb{R}^{N \times M}$, and $\bm{F}_{t-1}$ is the set of feedbacks given before the current query iteration. The summation in $n$ runs over the training data. The $u_{n,d,m,t}$ measure the influence that the feedback on feature $m$ would have for the predictive distribution of the $n$th patient for drug $d$. Once the query $(d^*,m^*)$ is selected and presented to the expert, the provided feedback is added to the set $\bm{F}_{t-1}$ to produce $\bm{F}_{t}$. Queries where the expert is not able to provide an answer do not affect the prediction model, but are added to the set so as not to be repeated.

Using the approximated posterior distribution (see Appendix A), the posterior predictive distribution of the relevance and directional feedback,\newline $p(\tilde{f}^{rel}_{d,m},\tilde{f}^{dir}_{d,m}|\mathcal{D}_{t-1}) = p(\tilde{f}^{rel}_{d,m}|\mathcal{D}_{t-1}) p(\tilde{f}^{dir}_{d,m}|\mathcal{D}_{t-1})$, follows a product of Bernoulli distributions. The approximate posterior predictive distribution of $\tilde{y}_d$ follows a Gaussian distribution which makes the $\KLt$ divergence calculation simple. Calculating the expected information gain for each (drug,~feature) pair requires four posterior approximations, which would make the query phase too costly. We follow a similar approach as in \cite{Daee2016}, and approximate the posterior with the new feedbacks with only partial expectation propagation updates. 

\subsubsection{User model} \label{user_model}
We introduce another approach for selecting the next (drug,~feature) pair candidate using a multi-armed bandit user model, extending the work in \cite{Micallef2017}. The benefit of bandit user modeling is that the model learns from the previous answers of the expert, and can guide the elicitation towards (drug,~feature) pairs that will most likely get an answer from the expert (\textit{exploitation}), simultaneously balancing the trade-off with \textit{exploration} of  
uncertain pairs.
We borrow this idea from the bandit literature (see, for instance,~\cite{LaiRobbins1985}) to ensure that our user model concentrates the queries to the (drug,~feature) pairs that are likely to get an answer from the expert.

The user model predicts the expected response of the expert for each query, in order to select the query on which we ask feedback next. We follow a linear bandit model~\cite{auer2002using} and previous work on user intent modeling~\cite{ruotsalo15interactive, Micallef2017}, where the estimate for each query is given by a dot product between a feature vector describing the query (later \textit{description vector}, $\mathbf{z}_{d,m}$), and an unknown parameter that gives the relevance of the queries. The expected response is then $E[r_{d,m}] = \mathbf{z}_{d,m}^{\top}\mathbf{v}$, and $\mathbf{v}$ at the iteration $t$ is estimated using standard linear regression $\hat{\mathbf{v}}_t = (\mathbf{Z}_t^\top \mathbf{Z}_t + \lambda \mathbf{I})^{-1} \mathbf{Z}_t^\top(\mathbf{r}_t-b)$.
Here $\lambda$ is a regularizer, $\mathbf{Z}_t$ is a matrix containing description vectors $\mathbf{z}_{d,m}$ of the pairs $(d,m)$ that have received feedback before or at the iteration $t$, and similarly $\mathbf{r}_t$ contains the $t$ responses of the expert before or at the iteration $t$. The response of the expert is $r_{d,m}=1$ if the feedback for the pair $(d,m)$ is either "relevant", directional or "irrelevant", and $r_{d,m}=0$ if the answer is "I don't know". The default response $b$ is set to 0.5. The model chooses the pair for the next query based on the upper confidence bound criterion \cite{auer2002using, chu2011contextual}. The details of the user model are given in the Appendix B.

A simple, common choice for the description vector would be to use directly the patient measurements (as done in previous works, for instance, for news article recommendation, where the description vector corresponding to each news was given by the existing features in the news article dataset and the user features~\cite{Li2010www}). However, in the more difficult case of precision medicine, this simple description vector definition would not lead to good performance due to the small sample size. Furthermore, previous studies show that the use of auxiliary data is effective in both drug response prediction \cite{costello2014community} and interactive expert knowledge elicitation \cite{Micallef2017}. Thus, we introduce description vectors for each (drug $d$,~feature $m$) pair, constructed by using prior knowledge in the form of KEGG pathways from Molecular Signatures Database (MSigDB)~\cite{liberzon2011molecular}, and the drug target genes from the DrugBank \cite{wishart2006drugbank}. 

Specifically,  
we first indicate if the feature is the target of the drug, and then if the feature belongs in the same pathway as the target of the drug. The description of mutation features is included as an indication of which pathways the mutated gene belongs to. In our experiments (see Section~\ref{sec:experiments}), this results in description vectors of length 133, including 1 description of the feature type (mutation or a cytogenetic marker), 2 descriptions of the (drug,~feature) pair, and 130 KEGG pathways containing any of the included genes, specified in Section~\ref{feedback_collection}.

\section{Experiments}\label{sec:experiments}

In order to evaluate the proposed methods, we apply them to real patient data and use feedback from well-informed experts to simulate sequential knowledge elicitation. Details of the data set and the expert feedback collection are presented in the next section, followed by the experimental results showing the effectiveness of the methods in practice.

\subsection{Experimental methods}\label{sec:exp_methods}
We used a complete set of measurements on \textit{ex vivo} drug response, somatic mutations and karyotype data (cytogenetic markers), generated for a cohort of 51 multiple myeloma patient samples. Drug responses are presented as Drug sensitivity scores (DSS) as described in \cite{yadav2014quantitative} and were calculated for 308 drugs that have been tested in 5 different concentrations over a 1000-fold concentration range. Somatic mutations were identified from exome sequencing data and annotated as described earlier in \cite{kontro2014novel}.

We focus our analysis on 12 targeted drugs, grouped in 4 groups based on their primary targets (BCL-2, Glucocorticoid, PI3K/mTOR, and MEK1/2). This results in data matrices of $51\times12$ (samples vs. drugs), $51\times3025$ (samples vs. mutations), and $51\times7$ (samples vs. cytogenetic markers). All data are normalized to have zero mean and unit variance. In this paper we ask the experts only about the somatic mutations and cytogenetics markers, which the experts know better and hence need to spend less time on in the experiments. We will extend to molecular features with less well known effects, such as gene expression, in follow-up work.

We use leave-one-out cross-validation\footnote{That is, in computing the predictions for each patient, that particular patient is not used in learning the prediction model.} to estimate the performances of the drug sensitivity prediction models, with the concordance index (C-index; the probability of predicting the correct order for a pair of samples; higher is better)\footnote{We note that C-index computed from leave-one-out cross-validation can be biased as it compares predictions for pairs of samples. We do not expect this to favour any particular method.} \cite{harrell2015regression, costello2014community} and the mean squared error (MSE; lower is better) as the performance measures. MSE values are given in the normalized DSS units (zero mean, unit variance scaling on training data). Bayesian bootstrap \cite{rubin1981bayesian} over the predictions is used to evaluate the uncertainty in pairwise model comparisons: in particular, we compute the probability that model $M_1$ is better than model $M_2$ as $\Pr(M_1 \text{ is better than } M_2) = \frac{1}{B} \sum_{b=1}^B I(M_1 \text{ is better than } M_2 \text{ in bootstrap sample } b)$, where $I(C)=1$ if condition $C$ holds and $0$ otherwise \cite{vehtari2002bayesian}.

The hyperparameters of the prediction model were set as $\alpha_\sigma=4$, $\beta_\sigma=4$, $\alpha_\rho=1$, $\beta_\rho=2$, $\mu=-2$, $\omega^2=\frac{1}{2}$, and $\alpha_\pi=9$, $\beta_\pi=1$ to reflect relatively vague information on the residual variance (roughly higher than $0.5$), a minor preference for sparse models and moderate effect sizes, and the \textit{a priori} quality of the expert knowledge as 9 correct feedbacks out of 10. The regularization in the user model is $\lambda=10^{-3}$, following the work~\cite{Micallef2017}.

\subsection{Feedback collection} \label{feedback_collection}

We expect that the experts can give feedback on the relevance and the direction of the putative effect of features in predicting the response to a drug. As observed in practice, the effect of a feature is often specific to a certain type of drug, 
therefore, we decided to elicit feedback on (drug,~feature) pairs. Furthermore, we note that the experts indicated that the same feedback 
applies to all drugs in the same drug group. 
Specifically, we collected feedback from two well-informed experts of multiple myeloma, using a form containing 161 mutations known to be related to cancer~\cite{COSMIC}, and 7 cytogenetic markers. The experts were asked to give feedback specific to 12 targeted drugs, grouped by the targets (BCL-2, Glucocorticoid, PI3K/mTOR, and MEK1/2).
The answer counts by feedback type can be found in the Appendix C.
The experts were instructed not to refer to external databases while completing the feedback form, in order to collect their (tacit) prior knowledge on the problem and make the task faster for them.

\subsection{Simulated user experiment} \label{user_experiment}
We simulate sequential expert knowledge elicitation by iteratively querying (drug,~feature) pairs for feedback, and answering the queries using the pre-collected feedback described in Section~\ref{feedback_collection}. At each iteration, the models are updated and the next pair is chosen, based on the feedback elicited up to that iteration, and the measurement data set which does not change. We run three simulations for comparing the elicitation methods: two where the pairs are chosen using one of the methods presented in Sections~\ref{information_gain} and~\ref{user_model}, and one where the pair is chosen randomly. The pairs are selected without replacement from the 2016 ($=12 \times 168$) pairs included in the feedback collection. The rest 2864 mutations are not queried for feedback, but all 3025 are included in the prediction model.

\subsection{Results}
We present here the two main results of the experiments. Further supplementary results can be found in Appendix D.

\vspace{0.08in}
\noindent \textbf{Expert knowledge elicitation improves the accuracy of drug sensitivity prediction.}
Table~\ref{Tab:no_feedback_performances} establishes the baselines by comparing the prediction model we use, the spike-and-slab regression model without expert feedback, to constant prediction of training data mean, ridge regression, and elastic net regression\footnote{Ridge regression and elastic net are implemented using the glmnet R-package \cite{friedman2010} with nested cross-validation for choosing the regularization parameters.}. Elastic net has poor performance with regard to MSE on this dataset. The ridge and the spike-and-slab models have comparable performances, with bootstrapped probabilities of 0.87 of ridge being better in the C-index and 0.42 in MSE.

Table~\ref{Tab:feedback_performances} compares the spike-and-slab model without feedback to the models with all expert feedback. Knowledge of both of the experts improves the predictions. The model with feedback from the senior researcher has 7\% higher C-index and 8\% lower MSE compared to the no feedback model and is confidently better according to the bootstrapped probabilities (0.98 for C-index and 0.95 for MSE).
The predictions improve in all of the 12 drugs that were considered in the experiment. Detailed results of drug-wise predictions are provided in Appendix D.

\begin{table}[htbp]
  \centering 
  \caption{Performance of drug sensitivity prediction without expert feedback. Values are averaged over the 12 drugs. Best result on each row has been boldfaced.} 
  \begin{tabular}{@{}lrrrr@{}}\hline
    & Data mean & Ridge & Elastic net & Spike-and-slab \\ \hline
    C-index & 0.50 & \textbf{0.62} & 0.60 & 0.61 \\  
    MSE & 1.06 & 0.94 & 1.00 & \textbf{0.93} \\ \hline
  \end{tabular}
  \label{Tab:no_feedback_performances}
\end{table}

\begin{table}[htbp]
  \centering 
  \caption{Predictive performance of spike-and-slab regression with and without expert feedback. Values are averaged over the 12 drugs.} 
  \begin{tabular}{@{}lrrr@{}}\hline
     & No feedback & Doctoral candidate & Senior researcher \\ \hline
    C-index & 0.61 & 0.63 & \textbf{0.65} \\  
    MSE & 0.93 & 0.92 & \textbf{0.86} \\ \hline
  \end{tabular}
  \label{Tab:feedback_performances}
\end{table}

\begin{figure*}
\centering
\vspace{-0.9in}
    \subfigure {\label{fig:MSE_norm_drug_feature_SR}
    \includegraphics[width=0.45\textwidth,keepaspectratio]{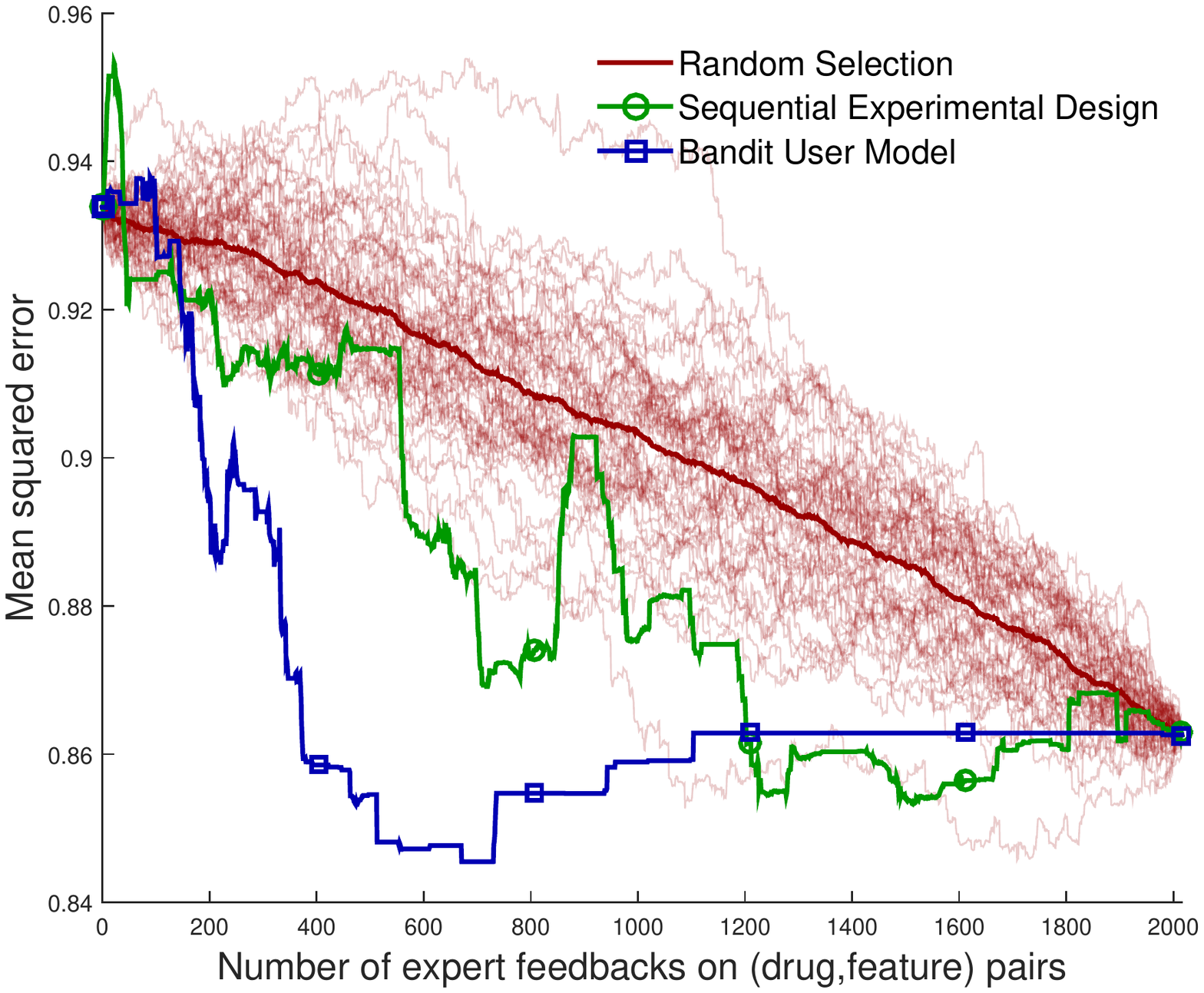}
    }
    \subfigure {\label{fig:MSE_norm_drug_feature_DC}
    \includegraphics[width=0.45\textwidth,keepaspectratio]{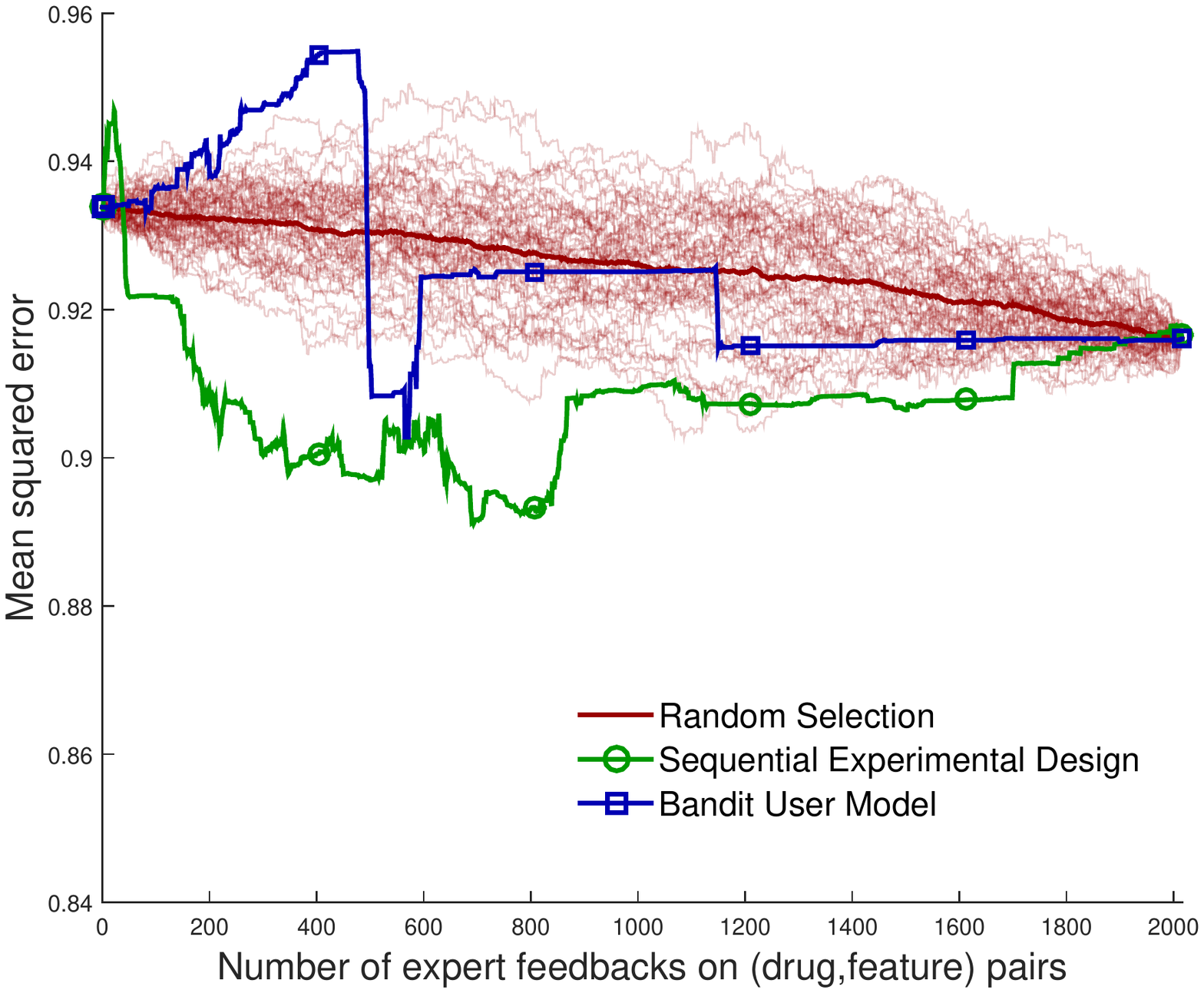}
    }
\caption{Performance improves faster with the active elicitation methods than with randomly selected feedback queries. The curves show mean squared errors as a function of the number of iterations for the three query methods, with feedback of the senior researcher (left) and doctoral candidate (right). In each iteration a (drug,~feature) pair is queried from the expert. The 50 independent runs of randomly selected queries are shown in red, and the average in thick red line.}
\label{fig:MSE_norm_drug_feature}
\end{figure*}

\noindent \textbf{Sequential knowledge elicitation reduces the number of queries required from the expert.}
In the results presented so far, the experts had evaluated all 
(drug,~feature) pairs and given their answers. However, sequential knowledge elicitation has the potential to reduce this workload significantly. We compare the effectiveness of the elicitation methods developed in this paper using a simulated user experiment (see Section \ref{user_experiment}). The results in Figure~\ref{fig:MSE_norm_drug_feature} show that both methods achieve faster improvement in prediction accuracy than the random selection, as a function of the amount of feedback. We use Area Under the MSE Curves (AUC) to evaluate the significance of the improvements in predictions of our knowledge elicitation methods, compared to the random selection method\footnote{We compare the AUC values of our knowledge elicitation methods to the empirical distribution of the AUC values of 50 independent runs of the random selection method.}. With the senior researcher feedback, $p \leq 0.02$ for both methods
, whereas with the doctoral candidate feedback $p \leq 0.02$ for the sequential experimental design and $p = 0.35$ for the bandit user model. With sequential knowledge elicitation, 50 \% of the final improvement is reached in the first 496 (43) and 191 (562) feedbacks, for the experimental design and bandit user model respectively, using Senior researcher feedback (Doctoral candidate feedback). For a comparison, an average of 1139 (1228) feedbacks are required for similar accuracy if the queries are chosen randomly. Thus, on average, the sequential experimental design requires only 23\% of the number of queries compared to random, and the bandit user model 32\%, to achieve half of the potential improvement.

\vspace{-0.1in}
\section{Conclusion}

In this work we show, for the first time, that sequential expert knowledge elicitation improves drug sensitivity prediction in precision cancer medicine. We also show in a simulated user experiment with real expert feedback that the proposed algorithms can elicit knowledge from experts efficiently. The results indicate that expert knowledge can be very beneficial and, hence, should be taken into account in modeling tasks of precision medicine. The doctors and researchers are analyzing the data regardless of the advances in the automated methods, and to not take their knowledge and expertise into account is to neglect one possible source of data in a case where the lack of data is a significant problem.

Our results were based on on knowledge elicited from two experts only. Nevertheless, a significant improvement in knowledge elicitation performance was observed for each of them even individually. 
In the future we will carry out a wider study to thoroughly quantify the effect of expert feedback, and to investigate further the initial observations about the impact of the type of feedback and the level of seniority of the experts. We note that the experts in this study had seen the data before, but to minimize the risk of overfitting they were instructed to answer based on knowledge without seeing the data, and in the follow-up work we will recruit experts who are completely naive as to the particular data.

We found that the most efficient elicitation method was different for the two experts. An obvious next question is how to combine the two elicitation methods to optimally utilize the complementary principles in them. In addition, we have shown here the improvement in sparse linear regression models. The next step will be to extend the method to more complex nonlinear models, and study how to maximally benefit from the responses of multiple experts.


\section*{Acknowledgements}
This work was supported by the Academy of Finland [grant numbers  295503, 294238, 292334, 286607, 294015] and Centre of Excellence in Computational Inference Research COIN; and by Jenny and Antti Wihuri Foundation. We acknowledge the computational resources provided by the Aalto Science-IT project.

\bibliographystyle{ieeetr}
\bibliography{references}

\newpage

\section*{Appendix A. Prediction model} 

\begin{figure}
\centering
\includegraphics[width=0.45\textwidth,keepaspectratio]{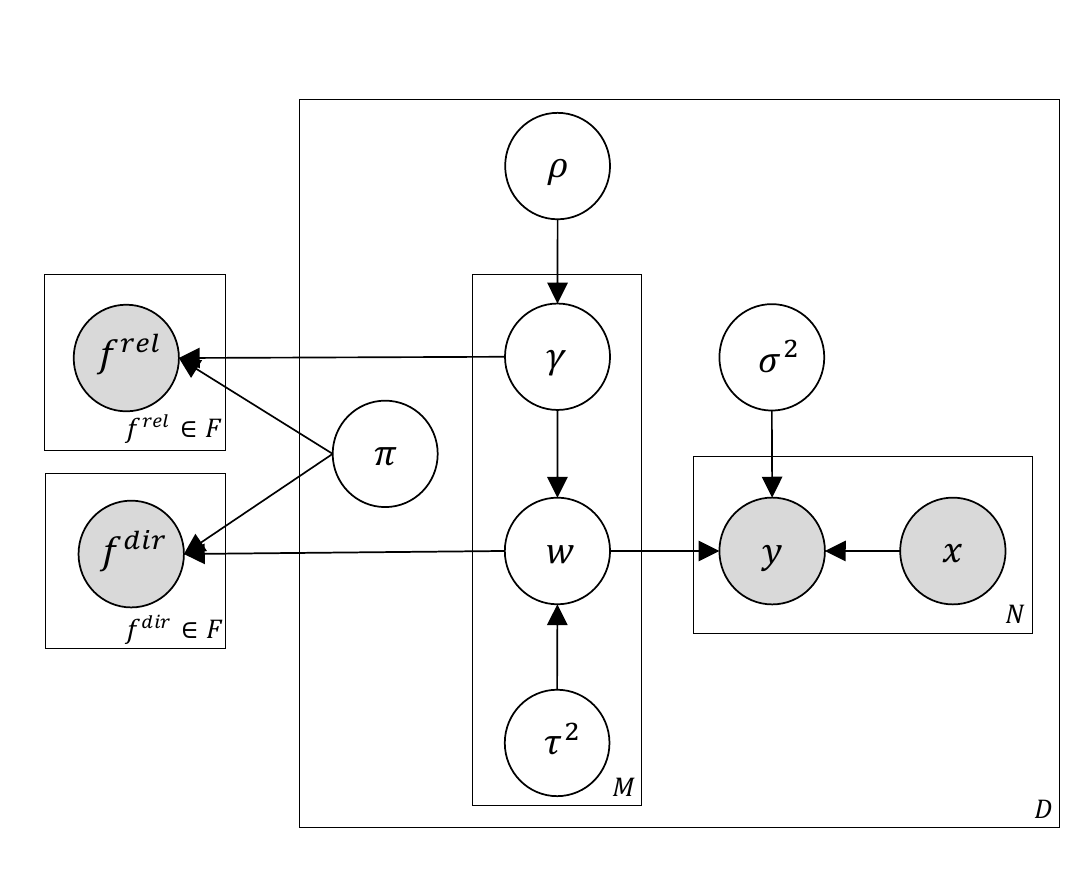}
\caption{Plate notation of the prediction model (right) and feedback observations (left) as introduced in Section~\ref{Prediction model} and Appendix A. The feedbacks $f^{rel}$ and $f^{dir}$ are sequentially queried from the expert based on an expert knowledge elicitation method.}
\label{fig:plate_diagram}
\end{figure}

Sparse linear regression models are used to predict the drug sensitivities based on the genomic features. Let $y_{n,d}$ be the sensitivity of the $n$th patient for drug $d$, and $\bm{x}_n \in \mathbb{R}^M$ be the vector of the patient's $M$ genomic features. We assume a Gaussian observation model:
\begin{equation*}
y_{n,d} \sim \normalpdf(\bm{w}_d\tp \bm{x}_n, \sigma^2_d),
\end{equation*}
where the $\bm{w}_d \in \mathbb{R}^M$ are the regression weights and $\sigma^2_d$ is the residual variance. A sparsity-inducing spike-and-slab prior \cite{mitchell1988bayesian, george1993variable} is put on the weights:
\begin{equation*}
w_{d,m} \sim \gamma_{d,m} \normalpdf(0, \tau_{d,m}^2) + (1 - \gamma_{d,m}) \delta_0,
\end{equation*}
where $\gamma_{d,m}$ is a binary variable indicating whether the $m$th feature is relevant (i.e., $w_{d,m}$ drawn from a zero-mean Gaussian prior with variance $\tau_{d,m}^2$) or not ($w_{d,m}$ is set to zero via the Dirac delta spike $\delta_0$) when predicting for the $d$th drug. The prior probability of relevance $\rho_d$ controls the expected sparsity of the model via the prior
\begin{equation*}
\gamma_{d,m} \sim \bernoullipdf(\rho_d).
\end{equation*}

The model is completed with the hyperpriors:
\begin{align*}
    \sigma^{-2}_d &\sim \gammapdf(\alpha_{\sigma}, \beta_{\sigma}),\\
    \rho_d &\sim \betapdf(\alpha_\rho, \beta_\rho),\\
    \tau_{d,m} &\sim \lognormalpdf(\mu, \omega^2).
\end{align*}
Settings for the values of the hyperparameters are discussed in Section~\ref{sec:exp_methods}.

Given the observed drug sensitivities $\bm{Y} \in \mathbb{R}^{N \times D}$ for $N$ patients and $D$ drugs and the genomic features $\bm{X} \in \mathbb{R}^{N \times M}$, the posterior distribution of the model parameters $\bm{\theta}=(\bm{w},\bm{\gamma},\bm{\rho},\bm{\tau}^2,\bm{\sigma}^2)$ is computed via the Bayes theorem as
\begin{equation*}
p(\bm{\theta} \mid \bm{Y}\!, \bm{X}) = \frac{p(\bm{Y}\!\mid\! \bm{X}, \bm{w}, \bm{\sigma^2}) p(\bm{w}\!\mid\!\bm{\gamma}, \bm{\tau^2}) p(\bm{\gamma}\!\mid\!\bm{\rho}) p(\bm{\rho}) p(\bm{\tau}^2) p(\bm{\sigma^2})}{p(\bm{Y} \mid \bm{X})}.
\end{equation*}
The posterior distribution of $\bm{w}$ together with the observation model is then used to compute the predictive distribution of the drug sensitivities $\tilde{\bm{y}} = [\tilde{y}_1,\ldots, \tilde{y}_D]\tp$ for a new data point $\tilde{\bm{x}}$:
\begin{equation}
p(\tilde{\bm{y}} \mid \bm{Y}, \bm{X}, \tilde{\bm{x}}) = \int p(\tilde{\bm{y}} \mid \tilde{\bm{x}}, \bm{w}, \bm{\sigma^2}) p(\bm{\theta} \mid \bm{Y}, \bm{X}) d{\bm{\theta}}.\label{Aeqn:preddist}
\end{equation}

Expert knowledge is incorporated into the model via feedback observation models \cite{Daee2016}. The relevance feedback $f^{rel}_{d,m} \in \{0,1\}$ ($0$ denotes irrelevant, $1$ relevant) of feature $m$ for drug $d$ follows
\begin{equation*}
f^{rel}_{d,m} \sim \gamma_{d,m} \bernoullipdf(\pi_d^{rel}) + (1 - \gamma_{d,m}) \bernoullipdf(1 - \pi_d^{rel}),
\end{equation*}
where $\pi_d^{rel}$ is the probability of the expert being correct. For example, when the $m$th feature for drug $d$ is relevant in the regression model (i.e., $\gamma_{d,m} = 1$), the expert would \textit{a priori} be assumed to say $f^{rel}_{d,m} = 1$ with probability $\pi_d^{rel}$. In the model learning (i.e., calculating the posterior distribution in Equation~\ref{Aeqn:posterior_with_feedback} below), once the expert has provided the feedback based on his or her knowledge, $\pi_d^{rel}$ effectively controls how strongly the model will change to reflect the feedback.

The directional feedback $f^{dir}_{d,m} \in \{0,1\}$ ($0$ denotes negative weight, $1$ positive) follows
\begin{equation*}
f^{dir}_{d,m}\!\sim\!I(w_{d,m}\!\geq\!0)\!\bernoullipdf(\pi_d^{dir})\!+\!I(w_{d,m}\!<\!0)\!\bernoullipdf(1\!-\!\pi_d^{dir}),
\end{equation*}
where $I(C) = 1$ when the condition $C$ holds and $0$ otherwise, and $\pi_d^{dir}$ is again the probability of the expert being correct. For example, when the weight $w_{d,m}$ is positive, the expert would \textit{a priori} be assumed to say $f^{dir}_{d,m} = 1$ with probability $\pi_d^{dir}$.

To simplify the model, we assume $\pi_d = \pi_d^{dir} = \pi_d^{rel}$ and set a prior on $\pi_d$ as
\begin{equation*}
 \pi_d \sim \betapdf(\alpha_\pi, \beta_\pi).
\end{equation*}

Given the data $\bm{Y}$ and $\bm{X}$ and a set of observed feedbacks $\bm{F}$ encoding the expert knowledge, the posterior distribution is computed as
\begin{align}
p(\bm{\theta} \mid \mathcal{D}) =& \frac{p(\bm{Y} \mid \bm{X}, \bm{w}, \bm{\sigma}^2) p(\bm{w} \mid \bm{\gamma}, \bm{\tau}^2) p(\bm{\gamma} \mid \bm{\rho}) p(\bm{\rho}) p(\bm{\tau}^2) p(\bm{\sigma}^2)}{p(\bm{Y}, \bm{F} \mid \bm{X})}\nonumber\\
& \times p(\bm{F} \mid \bm{\gamma}, \bm{w}, \bm{\pi}) p(\bm{\pi}),\label{Aeqn:posterior_with_feedback}
\end{align}
where $\mathcal{D} = (\bm{Y}, \bm{X}, \bm{F})$ and $\bm{\theta}$ now includes also $\bm{\pi}$. The predictive distribution follows from Equation~\ref{Aeqn:preddist}. Figure~\ref{fig:plate_diagram} shows the plate diagram of the model.

The computation of the posterior distribution is analytically intractable. The expectation propagation algorithm \cite{minka2002expectation} is used to compute an efficient approximation. In particular, the posterior approximation for the weights $\bm{w}$ is a multivariate Gaussian distribution and the predictive distribution for $\tilde{y}_d$ is also approximated as Gaussian \cite{Lobato2015ML, Daee2016}. The mean of the predictive distribution is used as the point prediction in the experimental evaluations in Section \ref{sec:experiments}. 

\section*{Appendix B. User model} 
The user model chooses the pair for the next query based on upper confidence bound criterion. The upper confidence bound at iteration $t$ is computed as $r^{UCB~(t)}_{d,m} = r_{d,m}^{(t)} + c_{d,m}^{(t)}$, where the confidence of the response is computed as in~\cite{chu2011contextual}:
\begin{equation*}
    c_{d,m}^{(t)} =\rho_t \sqrt{\mathbf{z}_{d,m}^\top(\mathbf{Z}_t\mathbf{Z}_t^\top + \lambda \mathbf{I})^{-1}\mathbf{z}_{d,m}}, \hspace{0.4cm}
     \rho_t = \sqrt{\frac{1}{2}\log \big(\frac{2tK}{\delta}\big)}. 
\end{equation*}
Here $K$ is the number of (drug,~feature) pairs, and $\delta=0.05$ defines that the bound holds with the probability $1-\delta$. The user model is initialized using regression weights from the prediction model as pseudo-feedback, with lower weight such that one feedback from an expert corresponds to 10 pseudo-feedbacks, similarly as in~\cite{Micallef2017}.

\section*{Appendix C. Feedback collection} 
The answer counts by feedback type are summarized in Table~\ref{Tab:feedback_count} for both of the experts.
\begin{table}[htbp]
  \centering 
  \caption{Feedback type and count, given to the 2016 (drug,~feature) pairs by the experts. SR = Senior researcher, DC = Doctoral candidate.} 
  \begin{tabular}{@{}lrr@{}}\hline
    Answer & SR & DC \\
        \hline
        Relevant, positive correlation & 192 & 47\\
        Relevant, negative correlation & 14 & 34\\
        Relevant, unknown correlation direction & 26  & 372\\
        Not relevant & 13 & 0\\
        I don't know  & 1771 & 1563\\ \hline
  \end{tabular}
  \label{Tab:feedback_count}
\end{table}

\section*{Appendix D. Further results} 
\vspace{0.08in}
\noindent \textbf{Feedback on the direction of the putative effect is more effective than general relevance feedback.}
We also assess the importance of the type of the feedback by comparing a spike-and-slab model with only relevance feedback (interpreting potential expert knowledge on the direction only as relevance) to a model with both types of feedback. Table~\ref{Tab:feedback_type_performances} shows that the directional feedback improves the performance markedly, especially in the case of the senior researcher (who gave more directional feedback than the doctoral candidate; see Table~\ref{Tab:feedback_count}). The bootstrapped probabilities are 0.70 in the C-index and 0.71 in MSE in favour of both types of feedback compared to only relevance feedback for the doctoral candidate and, similarly, 0.93 and 0.93 in the case of the senior researcher. 
\begin{table}[htbp]
  \centering 
  \caption{Performance of drug sensitivity prediction with only relevance feedback and with relevance and directional feedback. Values are averaged over the 12 drugs.} 
  \begin{tabular}{@{}lrr|rr@{}}\hline
     & \multicolumn{2}{c|}{Doctoral candidate} & \multicolumn{2}{c}{Senior researcher}\\ \hline
    & Relevance only & Both & Relevance only & Both \\ \hline 
    C-index & 0.62 & \textbf{0.63} & 0.62 & \textbf{0.65} \\
    MSE & 0.93 & \textbf{0.92} & 0.93 & \textbf{0.86} \\ \hline
  \end{tabular}
  \label{Tab:feedback_type_performances}
\end{table}

\vspace{0.08in}
\noindent \textbf{Use of biological prior information from databases benefits the user modeling.}
In the sequential knowledge elicitation with bandit user model, the queries are chosen based on expert's earlier answers and feature descriptions. However, the expert's feedback still defines which of the queried pairs are, in fact, relevant to the prediction task. Next we investigate how much the user models improve when using auxiliary biological information from databases. 
We form description vectors, as described in Section~\ref{user_model}, from the patient data used in the prediction task or, alternatively, from pathway and drug target information available in the databases. We compare the two alternatives in how well they discriminate the (drug,~feature) pairs for which the expert was able to provide feedback from those where the answer was 'I don't know'. The results in Table~\ref{Tab:classification_fd} show that using biological prior information improves especially the recall of the useful (drug,~feature) pairs, which means that the model finds a greater proportion of the (drug,~feature) pairs that got feedback.

\begin{table}[htbp]
  \centering 
  \caption{Precision and recall of classifying (drug,~feature) pairs using description vectors from patient sample data and from auxiliary data of pathways and drug targets. SR = Senior researcher, DC = Doctoral candidate} 
  \begin{tabular}{@{}llrr@{}}\hline
     &  & Patient data & Pathway and target information \\ \hline
    SR & Precision & \textbf{1.00} & 0.89 \\ 
     & Recall & 0.27 & \textbf{0.42} \\ \hline 
    DC & Precision & 0.88 & \textbf{0.90} \\
     & Recall & 0.23 & \textbf{0.46} \\ \hline
  \end{tabular}
  \label{Tab:classification_fd}
\end{table}

\vspace{0.08in}
\noindent \textbf{Drug-wise overview of the results.}
Table~\ref{Tab:individual_drugs} shows the effect of expert feedback on the 12 drugs used in the study. The predictions improved in all of the drugs, and the improvement is in general greater with the senior researcher's feedback. On the other hand, we observed that the type of feedback each expert gave was different, as the senior researcher provided more directional feedback than the doctoral candidate, and the doctoral candidate provided more relevance feedback (Table~\ref{Tab:feedback_count}). The greater number of directional feedback could explain the greater overall improvement with senior researcher's feedback, as we have already observed that directional feedback is more effective than relevance feedback.

\begin{table}[htbp]
  \centering 
  \caption{Predictive performance of individual drugs with and without expert feedback.\newline NF = No feedback, DC = Doctoral candidate, SR = Senior researcher.} 
  \begin{tabular}{@{}lrrr|rrr@{}}\hline
     & \multicolumn{3}{c|}{C-index} & \multicolumn{3}{c}{MSE}\\
    Drug & NF & DC & SR & NF & DC & SR\\
    \hline
    Pimasertib & 0.63 & 0.60 & \textbf{0.68} & 0.78 & 0.83 & \textbf{0.67}\\
    Refametinib & 0.67 & 0.64 & \textbf{0.68} & 0.80 & 0.85 & \textbf{0.71}\\
    Trametinib & 0.66 & 0.65 & \textbf{0.70} & 0.84 & 0.87 & \textbf{0.71}\\
    Dexamethasone & 0.65 & \textbf{0.71} & 0.68 & 0.96 & \textbf{0.91} & 0.96\\
    Methylprednisolone & 0.65 & \textbf{0.69} & 0.63 & 0.95 & \textbf{0.90} & 0.96\\
    AZD2014 & 0.61 & 0.60 & \textbf{0.68} & 0.94 & 0.93 & \textbf{0.80}\\
    Dactolisib & 0.59 & 0.59 & \textbf{0.66} & 0.97 & 0.92 & \textbf{0.86}\\
    Idelalisib & 0.45 & \textbf{0.55} & 0.52 & 1.12 & \textbf{1.09} & 1.25\\
    PF.04691502 & 0.57 & 0.62 & \textbf{0.64} & 1.00 & 0.96 & \textbf{0.90}\\
    Pictilisib & 0.57 & \textbf{0.64} & \textbf{0.64} & 0.95 & 0.88 & \textbf{0.87}\\
    Temsirolimus & 0.59 & 0.57 & \textbf{0.63} & 0.94 & 1.02 & \textbf{0.78}\\
    Venetoclax & 0.66 & \textbf{0.71} & 0.68 & 0.95 & \textbf{0.83} & 0.88\\ \hline
  \end{tabular}
  \label{Tab:individual_drugs}
\end{table}

\end{document}